\documentclass[preprint,12pt]{elsarticle}
\usepackage{changepage}
\usepackage{amssymb}
\usepackage{multirow}
\usepackage{float}
\usepackage{color} 
\journal{Nuclear Physics B}
\begin{document}
\begin{frontmatter}
\title{Bengali Sign Language Recognition through Hand Pose Estimation using Multi-Branch Spatial-Temporal Attention Model}



\author[inst1]{Abu Saleh Musa Miah}
\cortext[cor1]{Corresponding author}
\author[inst2]{Md. Al Mehedi Hasan}
\author[inst3]{Md Hadiuzzaman}
\author[inst4]{Muhammad Nazrul Islam}
\author[inst1]{Jungpil Shin\corref{cor1}}

\affiliation[inst1]{organization={School of Computer Science and Engineering,The University of Aizu},
            addressline={Address One}, 
            city={Aizuwakamatsu},
            postcode={965-8580}, 
            state={Fukushima},
            country={Japan.}}
        
\affiliation[inst2]{organization={Department of Computer Science and Engineering},
            addressline={Rajshahi University of Engineering and Technology(RUET)}, 
            city={Rajshahi},
            country={Bangladesh.}}
\affiliation[inst3]{organization={Department of Computer Science and Engineering, Bangladesh Army University of Science and Technology(BAUST)},
            addressline={Saidpur Cantonment, Saidpur}, 
            city={Nilphamari},
            postcode={Saidpur 5311}, 
            state={Rangpur},
            country={Bangladesh.}}
\affiliation[inst4]{organization={Department of Computer Science and Engineering (CSE), Military Institute of Science and Technology (MIST)},
            addressline={Mirpur Cantonment}, 
            city={Dhaka},
            postcode={Dhaka-1216}, 
            country={Bangladesh.}}
\begin{abstract}
Hand gesture-based sign language recognition (SLR) is one of the most advanced applications of machine learning, and computer vision uses hand gestures. However, many standard automatic recognition systems have been developed for other languages such as English, Turkish, Arabic, British etc. However, there still are some challenges to achieving standard Bangla sign language (BSL) recognition systems. Although, in the past few years, many researchers have widely explored and studied how to address BSL problems, certain unaddressed issues still remain, such as skeleton- and transformer-based BSL recognition. In addition, the lack of evaluation of the BSL model in various concealed environmental conditions can prove the generalized property of the existing model by facing daily life signs. As a consequence, existing BSL recognition systems provide a limited perspective of their generalisation ability as they are tested on datasets containing few BSL alphabets that have a wide disparity in gestures and are easy to differentiate. To overcome these limitations, we propose a spatial-temporal attention-based BSL recognition model considering hand joint skeletons extracted from the sequence of images. The main aim of utilising hand skeleton-based BSL data is to ensure the privacy and low-resolution sequence of images, which need minimum computational cost and low hardware configurations. Our model captures discriminative structural displacements and short-range dependency based on unified joint features projected onto high-dimensional feature space. Specifically, the use of Separable TCN combined with a powerful multi-head spatial-temporal attention architecture generated high-performance accuracy. The extensive experiments with a proposed dataset and two benchmark BSL datasets with a wide range of evaluations, such as intra- and inter-dataset evaluation settings, demonstrated that our proposed models achieve competitive performance with extremely low computational complexity and run faster than existing models.
\end{abstract}




\end{frontmatter}


\section{Introduction}
\label{sec1}
Language is the primary medium through which one can express thoughts, ideas, and requirements to other people. Some groups are deprived of using language to express their thoughts or requirements or hear something from another person. These people are usually not able to establish communication through the common language. They use other ways to fulfil their requirements. In our society, we know them as deaf and hard-of-hearing people who are mainly incapable of speaking using oral language \cite{shin2024japanese_jsl1,kakizaki2024dynamic_jsl2,shin2023korean_ksl1,10360810_miah_ksl2,shin2024korean_ksl0}. In some cases, they are not completely incapable of speaking but do not like to speak because of the negative voice or undesired attention that is possible to attract a typical voice. This deafness mainly comes from birth or hearing injury, yielding to complete inability to hear or moderate inability~\cite{kushalnagar2019deafness,cheok2019review}. In addition, hearing loss or deafness is associated with ageing, noise exposure, genetics and different kinds of infections such as chronic ear, certain toxins and medications~\cite{kushalnagar2019deafness,rahim2020hand,islam2024multilingual,}. According to a report from the World Health Organization (WHO), there are 1.33 billion people belonging to the deaf-mute community in the world, which is 18.5\% of the world population \cite{vos2016global} and, there are 300000 Bangladeshi people belonging to the deaf-mute community~\cite{alauddin2004deafness,miah2022bensignnet}. There are some treatments for hearing loss or deaf-mute problems if the deaf person cannot hear 25 decibels after the test ~\cite{kushalnagar2019deafness}. However, most of the people in the deaf-mute community are not suitable for the treatment due to high decibels. In this situation, there are some alternative solutions for the deaf and mute community, such as hearing aids, augmentative communication devices, cochlear implants, subtitles, and sign language ~\cite{kushalnagar2019deafness,miah2023dynamic_mcsoc,miah2023skeleton_euvip}. 
\par
In such a situation, establishing communication using a special gesture known as sign language comes from utilized visual-manual systems such as body movement and hand movement, which convey meaningful information to interact between the deaf or mute community~\cite{mindess2014reading,hassan2024deep_har_miah}. Although deaf people can generally learn sign language to establish communication among themselves, people are not interested in learning sign language to communicate with a specific group of people, like a deaf community. From this perspective, general people can get help from human interpreters. The scary and alarming side of the human interpreter is that it is costly and not easy to get expert human interpreters all the time when it is needed. As a consequence, the deaf and mute communities have been facing critical challenges in fulfilling their basic needs like medical series, socialization, education and employment interview process, etc. Besides sign language, there has been a growing demand for non-contact input interfaces in recent years because of COVID-19 and emergencies.  The whole world focused on using touchless devices for every sector that allow users to input data without touching their hands to avoid various infections. \par

Therefore, one of the aims of the study is to use a webcam or a single camera as an input device for communication. It is very hygienic and is expected to reduce the risk of infections. The advantage of using a webcam is that there is no need to prepare the above expensive devices. Many laptops already have a camera installed, so there is no need to buy additional devices. To deal with the following challenges, sign language and hand gestures \cite{mallik2024virtual,rahim2024advanced_miah}. Because of these reasons, automatic hand gestures and sign language recognition have become an important medium for both the deaf community and general people as a means of communication. Many researchers have been working to develop such automatic recognition systems by considering vision and sensor-based systems. Many researchers proposed sensor-based sign language recognition systems, among them hand gloves~\cite{chouhan2014smart,assaleh2012low,shukor2015new}, leap motion ~\cite{lai2012gesture}, kinematic \cite{almeida2014feature} and accelerometers are most of them \cite{hongo2000focus}. Although a sensor-based system carries the efficiency of the skeleton data with limited movement, it needs a specialized device that is portable and has costly problems. 
\par
To solve the portability of the system, researchers focus on the vision-based system considering digital camera \cite{mohandes2004automation}, RGB depth camera ~\cite{podder2022bangla}, web camera, stereo camera, or 3D camera~\cite{awan2021improved}. Most of the sign language recognition systems have been developed using the vision-based approach with various machine learning and deep learning-based methodology~\cite{islam2018ishara,hoque2016automated,rafi2019image}. Although many studies developed for another sign language, few models have been developed for BSL recognition because of the inadequacy of the freely available BSL benchmark dataset. Most of the existing models for the BSL recognition model have been evaluated with their own dataset. There may be difficulties in producing good performance with the versatile environmental test dataset~\cite{islam2018ishara,hoque2016automated,rafi2019image}. There are few benchmark datasets available for BSL recognition \cite{islam2018ishara,hoque2016automated,rafi2019image,kubdsl}. Rafi et al. created a benchmark BSL dataset named 38 BdSL dataset, which contains 12,160 samples; after evaluating with their deep learning algorithm, they achieved 89.60\% accuracy \cite{rafi2019image}. Ishara-Lipi is another benchmark dataset for BSL recognition ~\cite{islam2018ishara}. Abedin et al. proposed a concatenated CNN model to improve the performance of the BSL recognition system with the BdSL dataset, and they achieved 91.50\% ~\cite{abedin2021bangla}. Although they applied different deep learning-based models,, the performance accuracy and efficiency of the systems may not be satisfactory for real-time implementation. To overcome the problems, Miah et al. employed a CNN-based BensignNet to improve the accuracy of the BSL recognition \cite{miah2022bensignnet}. They achieved 94.00\% for the BdSL dataset and 98.20\% for the KU-BdSL dataset. The main drawback of these systems is computation complexity, the complexity of the background and the lack of generalization. To overcome the generalization problems, Youme et al. proposed a CNN-based algorithm by including intra and inter-dataset evaluation settings in BSL recognition to increase the generalization property \cite{youme2021generalization}. However these methods still face computation complexity problems and various other issues in achieving high accuracy for BSL recognition tasks like background noise, self-occlusion, partial occultation, redundant background, and viewpoint of light variation and variety of illuminations. By considering the challenges recently, many researchers employed the skeleton-based spatial-temporal attention model to recognize sign language and hand gestures~\cite{chen2019construct,shi2020decoupled,miah2023dynamic}. \par However, (i) the computational complexity of the existing attention-based model is so high, (ii) we did not find any skeleton-based model for the BSL recognition task, (iii) No existing methods of the BSL recognition focus on improving the generalization property except one. Although skeleton-based sign language data does not contain appearance information, it is robust to variations in viewpoint, lighting conditions, and backgrounds, making it the most suitable approach but not compromising privacy. To overcome the three challenges, we proposed a skeleton-based BSL recognition system using spatial-temporal multi-head attention followed by the Separable temporal convolutional network (Sep-TCN) to encode the robust spatiotemporal features. The main contribution of the proposed mode is given below:
\begin{itemize}
    \item New Dataset: We created a new benchmark dataset, namely BAUST-BSL-38, to overcome the data lacking problem of the BSL domain. 
    \item Cost-effectiveness: We proposed an efficient BSL recognition system by extracting 2D skeleton key points from common images aiming to use low-specification cameras for monitoring. Experimental results proved the system's efficiency across multiple datasets with different evaluation settings. To our knowledge, this is the first work to evaluate the BSL recognition system and other evaluations where spatiotemporal attention-based BSL is employed.
    
     \item Time efficiency: As we proposed here, a SepTCN with the multi-branch of multi-head spatiotemporal attention-based feature aggregation with different attention-based networks to capture slow and fast spa-to-temporal articulations. Our model reduces the required number of parameters (1/2nd), which helps to reduce the computational complexity three times compared to existing architectures and the best-competing method. The extremely lower amount of the model parameter and the minimum floating point operations per second (FLOPS) proved that the proposed efficient system extracted powerful features without applying exorbitantly deep network layers as compared to other existing state-of-the-art spatial-temporal methods and without any noticeable drop in performance
     \item
    Generalization: Good performance with BSL recognition in evaluating inter and intra-dataset settings proved the proposed model's lead as better-generalized solutions.
\end{itemize}

This paper is organized as follows: Section \ref{sec2} provides the relevant literature review. Section \ref{sec3} describes the benchmark dataset of hand skeletons used to develop this work. Section \ref{sec4} describes the proposed multibranch spatial-temporal attention model. Section \ref{sec5} described the experimental results and different evaluation scenarios. Section VI concludes the paper, including some future work.

\section{Related works} \label{sec2}
Although much research has been done on different sign language recognition \cite{miah2024review,miah2024hand_multiculture,miah2024sign_largescale,computers12010013_multistage_musa,electronics12132841_miah_multistream_4,miah2024spatial_paa,shin2024korean_ksl0,shin2023korean_ksl1,10360810_miah_ksl2,shin2024japanese_jsl1,kakizaki2024dynamic_jsl2}, few are available for BSL recognition. Pixel and skeleton-based image recognition, hand pose estimation, and hand tracking recognition are mainly considered for sign language recognition. There are many researchers employed machine and deep learning algorithms for sign language recognition; among them, Aryanie et al. used Principal Component Analysis (PCA) for dimensionality reduction by including the K-Nearest Neighbor (KNN) algorithm as a classifier to recognize the American sign language~\cite{aryanie2015american}. 
Different researchers use support vector machine (SVM) to recognize different sign language based on local and global features, like 86\% achieved for Japanese sign language \cite{mukai2017japanese} and good performance achieved for Thai finger spelling~\cite{pariwat2017thai}. To classify Bangla sign language, many researchers used SVM, such as researchers \cite{hasan2015new} who achieved 96.46\% accuracy for 5400 images which is 6\% higher than the KNN algorithm; another research team achieved good accuracy with HOG features \cite{hasan2016machine}. Uddin et al. l achieved 97.70\% accuracy for the SVM with 2400 images based on the Gabor filter features, which is reduced with PCA algorithms ~\cite{uddin2016hand}. Yasir et al. applied PCAN with Linear discriminant analysis (LDA) to maximize inter-class distance and minimize intra-class scatter \cite{yasir2017bangla}. The same researchers extracted SIFT features from a vocabulary dataset and applied KNN and SVM to classify Bangla sign language (BSL). Although some researchers achieved good accuracy with machine learning algorithms, these systems face difficulties for large image or video datasets and most of the studies are based on the self-collection of small datasets, most of which are not available for comparative analysis.
\par
Currently, CNN-based methods have been prosperous for sign language recognition because of the efficiency and accurate processing of large-scale complex vision information or large-size video datasets. Ahmed et al. proposed an artificial neural network (ANN) to recognize BSL based on the fingertip position feature-based concept~\cite{ahmed2016bangladeshi}. Shafique et al. employed CNN-based deep learning architecture to recognize BSL recognition, where they recorded data from 25 different subjects and achieved good performance accuracy with different settings of evaluations \cite{islalm2019recognition}. Hoque et al. proposed a BSL recognition system using region-based CNN, and they evaluated the model with a dataset consisting of 1000 gestures and achieved around 98.00\% accuracy~\cite{bintey2018real}. Xception transformer employed by Urmee et al. to recognise BSL recognition based on 37 different gesture-based datasets and achieved around 98.00\% with their dataset~\cite{urmee2019real}.
Moreover, CNN was employed by different researchers to recognize BSL, such as a virtual reality-based hand-tracking controller, which reduced the 2\% error rate \cite{yasir2017bangla}, digits classification and achieved 92.00\% accuracy~\cite{islam2019ishara}. To solve the inadequate dataset problems, Sanzidul et al. created a new benchmark dataset, Ishara-Lipi, with 36 gestures of BSL recognition, and they achieved 92.74\% accuracy using a CNN model~\cite{rafi2019image}. Hosssain et al. employed a CNN-based capsule network to improve the model's performance and efficiency based on this dataset and achieved around 98.00\% accuracy \cite{islam2019ishara}. They employed a general CNN model to classify BSL and achieved higher accuracy after evaluating the Ishara-Lipi dataset compared to the previous model. Some researchers used different kinds of transformers for BSL recognition to improve the efficiency and performance of the BSL recognition systems, such as Densenet201 architecture and zero-shot learning (ZSL)~\cite{hasan2020classification}, and VGG16~\cite{nihal2021bangla}. Rafi et al. created a BSL benchmark dataset considering 38 alphabets, namely the BdSL-38 dataset, and after applying the VGG19 method, they achieved 89.60\% accuracy \cite{rafi2019image}. Also, they employed a transformer-based network, but their performance was unsatisfactory. Abedin et al. employed deep learning-based concatenated CNN to overcome the challenges and achieved 91.50\% accuracy with the previous 38BdSL dataset \cite{abedin2021bangla}. 
\par
Also, they improved the performance accuracy, but their generalization property is not good because of the few validation settings. To increase the generalization property, miah et al. employed a CNN based BenSignNet model where they achieved 94.00\% accuracy for the BdSL-38 dataset and good accuracy for the other two datasets~\cite{miah2022bensignnet}. Although they achieved good performance and claimed the generalization property because of the good accuracy of the three datasets, they did not evaluate the inter-dataset or cross-evaluation criteria. Youme et al. proposed a CNN-based algorithm by including intra and inter-dataset evaluation set-tings in BSL recognition to increase the generalization property \cite{youme2021generalization}. However, they worked to increase the generalization property of the system. Still, their computational complexity and efficiency are not good because they used pixel-based image information containing redundant background, light illumination, and partial occlusion issues. To overcome the background-related computational complexity problems more recently, researchers have focused on using a self-attention mechanism based on skeleton data instead of pixel-based information by reducing the long-range de-pendency~\cite{tian2018cr,liu2016spatio,vaswani2017attention}. The main concept of the attention-based model is to extract the relationship among the pixel or skeleton points. They consider three types of simultaneous matrix, namely Query, Key, and Values, then multiply the first and third matrix, divide it by the third matrix, and use the softmax function to make a weight matrix~\cite{dai2019transformer,shi2020decoupled}. Some researchers combined the attention with other CNN, RNN, and LSTM networks, but it increased the long-range dependencies, which is not the main goal of the attention model ~\cite{si2019attention,baradel2017human,song2017end}. 
\par
In addition, one researcher combined the Spatial, temporal attention model with the residual CNN and achieved 89.20\% and 93.60 accuracies for DHG and Shrec hand gesture datasets, respectively. Some researchers generated the idea of the Graph for the skeleton information and proposed a graph convolutional neural network(GCNN) to increase the efficiency and generalization property of the Sign language recognition system~\cite{yan2018spatial,si2018skeleton,chen2019construct}. Miah et al. proposed a multibranch of the Graph and general neural network to recognize hand gesture recognition by replacing actual graph structure with a deep learning-based structure. They achieved good accuracy for the gesture recognition for the DHG, SHREC and MSRA datasets. These methods also achieved good performance and efficiency in their model, but their computational complexity is still high. We proposed a skeleton-based BSL recognition system using a spatial-temporal attention model to overcome the various challenges. Where we used Sep-TCN before the spatial-temporal model to reduce the number of parameters of the model. To evaluate the model, we used the BSL dataset with inter- and intra-dataset evaluation settings to prove the system's generalization property. 

\section{Dataset}\label{sec3}

Bangla sign language (BSL) mainly comes from the American sign language (ASL) concept following the Bangla language. Although there are 39 consonants and 11 vowels in Bangla, researchers considered 38 alphabets among them as a BSL\cite{rafi2019image}. These 38 alphabets were selected by collaborating with various deaf-mute community organizations, foundations, and educational institutions. 
We experimented with three individual BSL datasets of this study, namely BAUST-BSL-38, 38 BdSL and KU-BdSL datasets described in Section \ref{sec3.1}, Section \ref{sec3.2}, and Section \ref{sec3.3}, respectively. The primary distinction between the BdSL dataset and the KU-BDSL dataset lies in the number of hand gesture signs they contain. Specifically, the BdSL dataset comprises 38 hand gesture signs used to represent the Bangla Sign Language Alphabet, while the KU-BDSL dataset includes 30 hand gesture signs. This disparity in the number of signs is a key differentiation between the two datasets.
In contrast, our newly created dataset also encompasses 38 hand gesture signs, aligning with the Bangla Sign Language alphabet, which mirrors the composition of the BDSL dataset. However, it is essential to highlight that the BDSL dataset has limitations in terms of the number of samples available, making it less suitable for comprehensive system experimentation and real-time deployment. To address this issue, we collected our proposed dataset from a deaf and dumb school, intentionally capturing data in diverse environments. This approach enhances the dataset's ability to evaluate systems effectively and enables real-life deployment, particularly benefiting the deaf and dumb community.

\subsection{BAUST-BSL-38 Dataset} \label{sec3.1}
\begin{figure}[H]
\centering
\centering 
\includegraphics[scale=0.30]{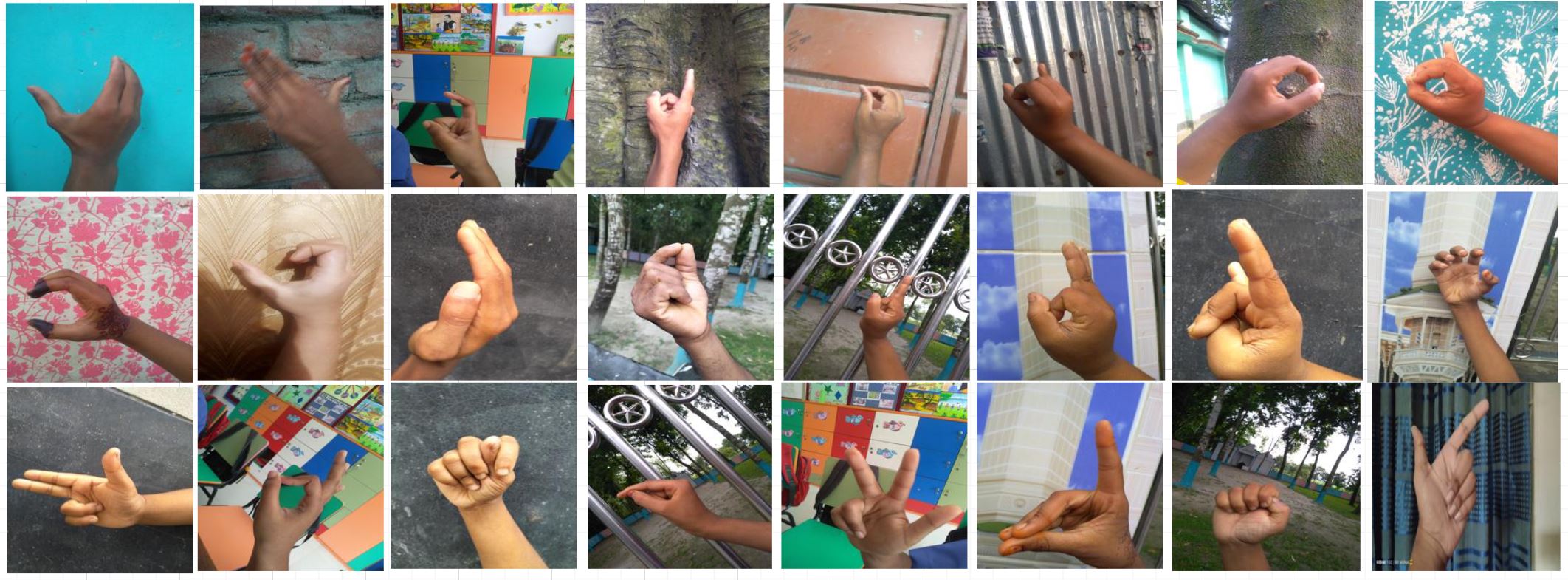}
\caption{Example of our collected dataset.}
\label{fig:Proposed_BdSL}
\end{figure}

Data acquisition is a crucial part of BSL recognition because it is not an efficient benchmark dataset for BSL recognition work~\cite{assaleh2012low,shukor2015new}. However, it is difficult to do this task because of the huge number of alphabets and various complexities in the BSL. To select the most usable alphabet, we used a BSL dictionary published by the National Centre for Special Education under the Ministry of Social Welfare, namely 'The Bangla Sign Language Dictionary'. To record the dataset, we collaborated with a deaf-mute organization, Proyash, Rangpur (A special education school). A sign language instructor of the proyash school, Rangpur, has instructed the subject to record our data collection. There are 38 standardized gestures for the 38 alphabets the national federation provides, shown in Figure \ref{fig:Proposed_BdSL}. The proposed dataset has been recorded from mixed people of general and hearing-impaired people. In total, 15 people gave the data; among them, 10 are normal people who are a student at the Bangladesh Army University of Science and Technology (BAUST), and 5 are deaf people who are a student at the Proyas School, Rangpur. Table \ref{Tab:Proposed_BdSL} describes the information of the people who participated in the data collection procedure. We recorded more than 600 samples for each of the 38 classes, and a total of around 22800 samples were collected. Multiple smartphone cameras have been used to collect the dataset; in most cases, it contains the single hand gesture dataset. Images were collected from the different complex backgrounds, and each image size was $512 \times 512$ pixels collected from different subjects. The sample image of the dataset is shown in Figure\ref{fig:Proposed_BdSL}.
\begin{table}[h]
\caption{Dataset Overview}
\label{Tab:Proposed_BdSL}
\setlength{\tabcolsep}{3pt}
\begin{tabular}{lllllll}
\hline
\multicolumn{2}{l}{Gender} & \multicolumn{2}{l}{Hearing Imparity} & Age Range & \multicolumn{2}{l}{Sample} \\ \hline
Male & Female & Deaf student & Normal student &  & Individual sign & Total \\
10 & 5 & 5 & 10 & 12-26 & 600 & 22800 \\ \hline
\end{tabular}
\end{table}

\subsection{BdSL-38 Dataset \cite{rafi2019image}} \label{sec3.2}
\begin{figure}[H]
\centering
\centering 
\includegraphics[scale=0.18]{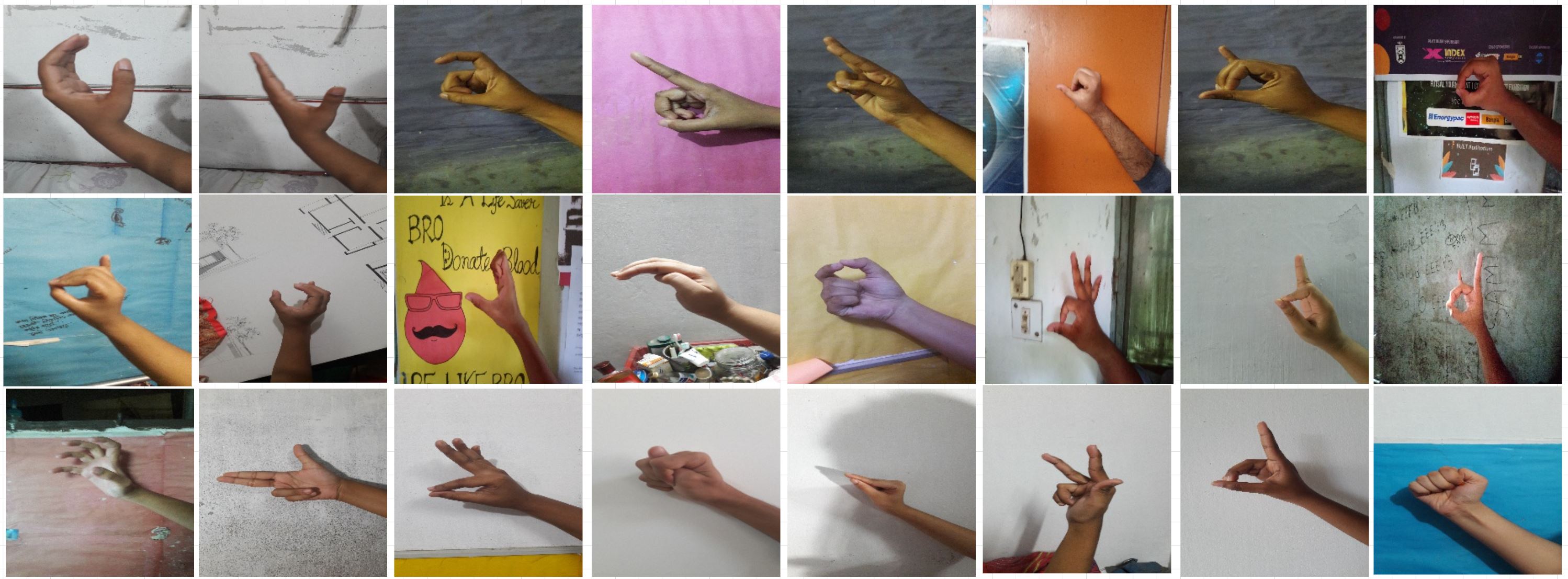}
\caption{Sample images of the 38 BdSL Dataset.}
\label{fig:dataset_BdSL-38}
\end{figure}

This is one of the most used benchmark datasets for BSL, which consists of 38 gestures of BSL \cite{rafi2019image}. This dataset was also recorded with the help of the National Federation of deaf people. They selected the 38 gestures by following the 'Bengali Sign Language Dictionary'. They recorded 320 images for each class, and in total, 12160 samples from the 320 people. Figure \ref{fig:dataset_BdSL-38} demonstrates the sample image of this dataset. 

\subsection{KU-BdSL \cite{kubdsl}} \label{sec3.3}
\begin{figure}[H]
\centering
\centering 
\includegraphics[scale=0.25]{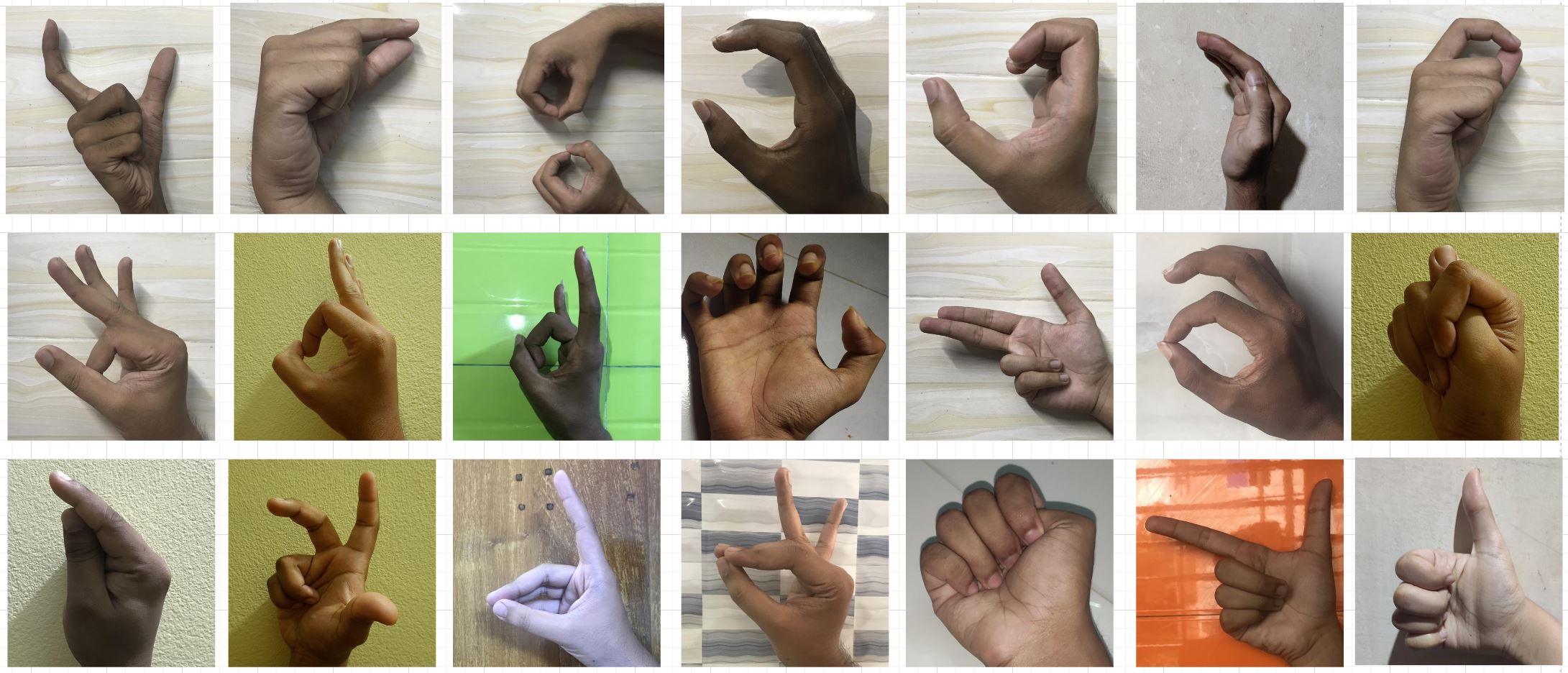}
\caption{Sample images of  the KU BdSL Dataset Images.}
\label{fig:KUBdSL}
\end{figure}

This is the other BSL dataset which name is the KU-BdSL Multi-scale sign language dataset \cite{kubdsl}. This dataset was recorded from 33 people, where 25 were male, and eight were female for 30 classes. Multiple smartphone cameras have been used to collect the dataset; in most cases, it contains the single hand gesture dataset. Images were collected from different complex backgrounds, and each image size was $512 \times 512$ pixels from different subjects. During the recording of the images, the hand image was placed in the middle position, which included versatile environmental conditions.
Figure \ref{fig:KUBdSL} demonstrates the sample image of the KU-BdSL dataset.

\section{Proposed Methodology}\label{sec4}

\begin{figure}[H]
\begin{adjustwidth}{-3cm}{-3cm}
\centering
\centering 
\includegraphics[scale=0.34]{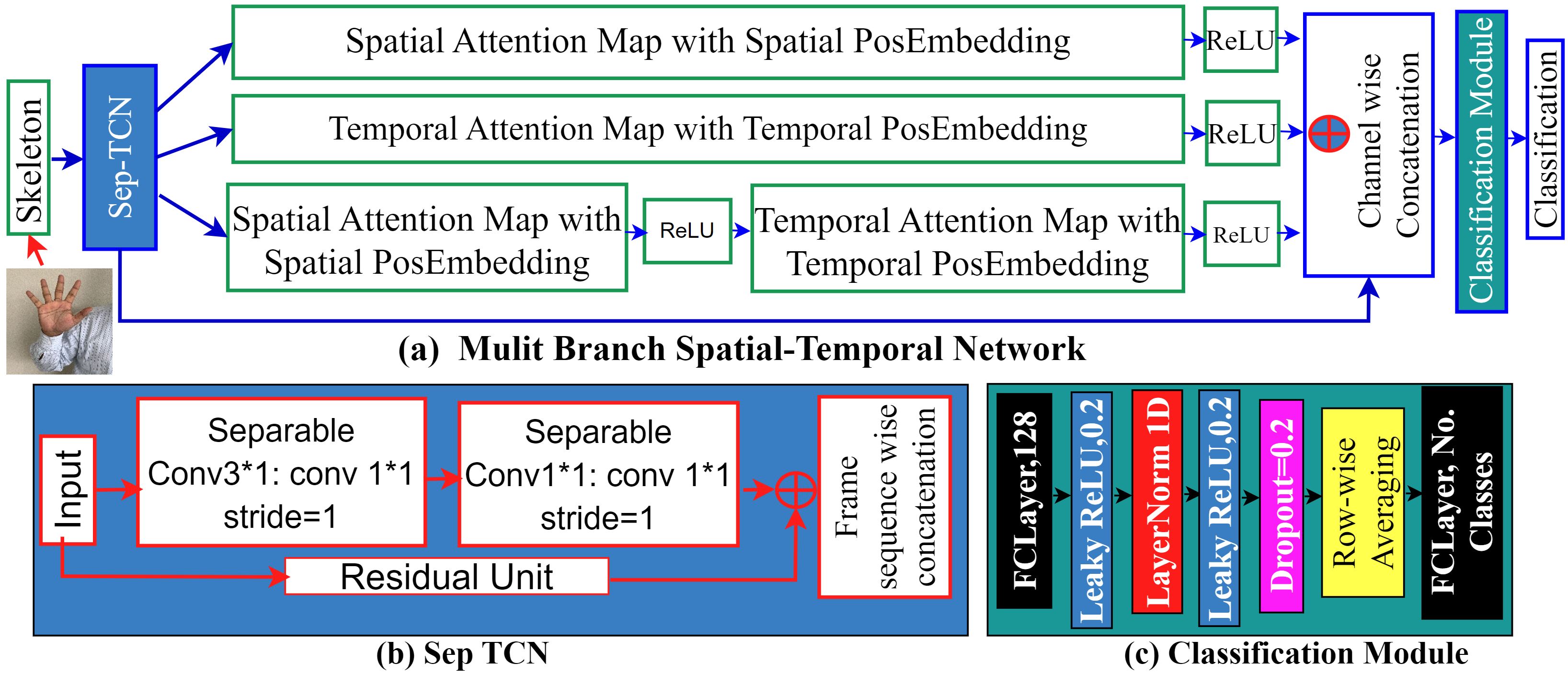}
\caption{Working Flow Architecture.}
\label{fig:main_diagram}
\end{adjustwidth}
\end{figure}

The working flow architecture of the proposed method is presented in Figure~\ref{fig:main_diagram}. Our main aim of this study is to develop a BSL recognition model to improve the accuracy and efficiency of real-life implementation. Our target is to reduce the computational complexity in order of magnitude lower computational complexity (in terms of FLOPS) with fewer parameters (1/2) and minimum (1/3) times compared to existing architectures compared to the best-competing method. To do this, we first collected the Bangla sign language dataset and then extracted the palm's joint key point from the image using a Media pipe. In the procedure, we extracted 2D 21 hand skeleton coordinates from BSL data images using a media pipe \cite{inproceedings,s21175856}. Then we combined the long, long-range and short-range dependencies to improve the performance of the BSL recognition. In the hand skeleton case, we employed an attention-based spatial and temporal network. Firstly, we employed a depth-wise separable convolutional network to extract effective features and convert the single image into a sequence of images. Then we parallelly employed the three branches of the attention-based architecture to extract effective features from the skeleton-based Bangla sign language dataset. 

\subsection{Hand Skeleton Estimation} \label{sec4.1}
\begin{figure}[H]
\centering
\centering 
\includegraphics[scale=0.30]{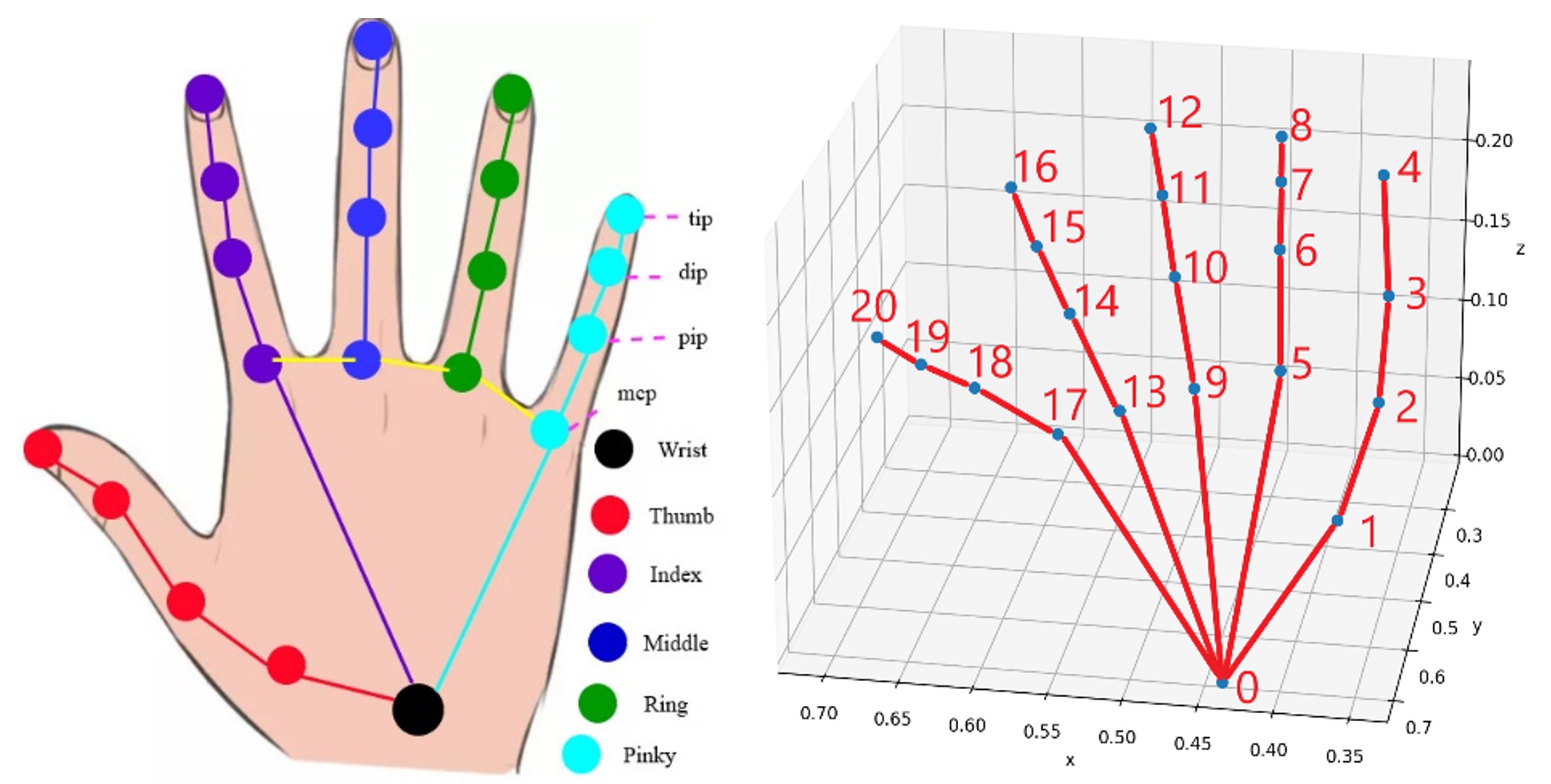}
\caption{Mediapipe key points.}
\label{fig:Keypoints}
\end{figure}

The RGB image of the hand obtained from the web camera is the input to the joint estimator to estimate the 3D coordinates of the hand joint. Media pipe Hands is a reliable hand and finger tracking API developed by Google to estimate the hand skeleton coordinates of each joint from a camera ~\cite{inproceedings}. It uses machine learning (ML) to understand 21 3D local hand marks from just one frame. The data obtained from the API consists of 21 points with x, y, and z coordinates. There is an order of these 21 coordinates as follows: The bottom point is the wrist which is the first coordinate, and from that, the thumb coordinates are in order from 1-4, namely thum cmc, thum mcp, thump ip and tip. After that, the index finger order is 5-8 from the wrist coordinate. In the same way, order 9-12 for the middle finger, 13-16 for the ring finger and 17-20 for the pinky fingers (see in Figure \ref{fig:Keypoints}). Although the position of each point is not fixed on either wrist or others, the coordinated value of each point usually changes with the movement of the hand. 

\subsection{Separable TCN} \label{sec4.2}
\begin{figure}[H]
\begin{adjustwidth}{-3cm}{-3cm}
\centering
\centering 
\includegraphics[scale=0.55]{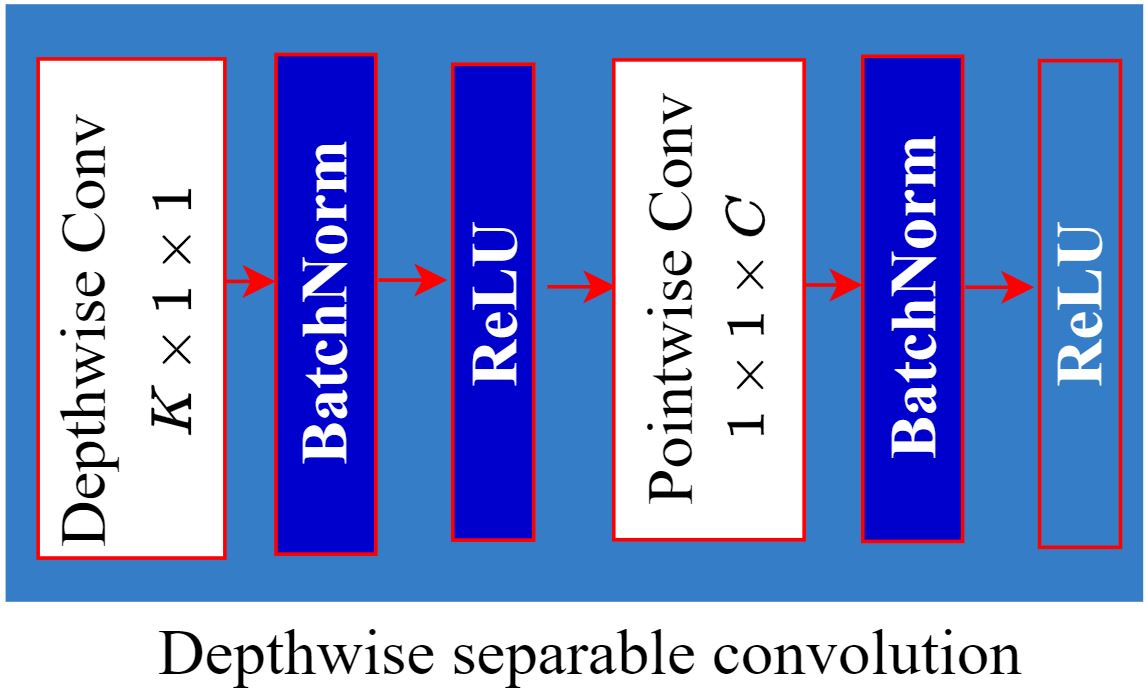}
\caption{Internal Structure of the separable TCN}
\label{fig:Sep_TCN}
\end{adjustwidth}
\end{figure}

A separable TCN, also known as a depthwise separable TCN, is a variant of the traditional Temporal Convolutional Network (TCN) architecture commonly used in deep learning for processing sequential data. The separable TCN is to reduce the number of parameters required by the traditional TCN while maintaining or improving its performance. In the traditional CNN, the kernel moves every time and needs K×K×C multiplication operations, exponentially increasing the total computation costs known as FLOPS. Separable TCN divided the full operation into two stages, namely depth-wise (DW) and pointwise (PW) convolution, which needs a minimum number of operations mathematically for each time DW needs K × K × 1, and PW needs 1 × 1 × C filter operation. This computation operation yields a sharply reduced number of operations for each kernel during the convolutions~\cite{howard2017mobilenets}. It also reduces the transpose operations in the convolution. Consequently, the total number of parameters of the proposed model is dramatically reduced without compromising the model's performance and efficiency, which means this makes the model faster twice compared to the traditional system and counterparts \cite{zahan2022sdfa}. In our study, for the first SepTCN layer, we used a 3 × 1 filter with stride one, and for the second Sep-TCN layer, we applied 5 × 1 with stride 2, yielding results with the filter size of 3×1×1 and 1×1×1 for DW. Figure \ref{fig:main_diagram}(b) demonstrated the filter size K × 1 × 1 where we did not compromise the original joint dimension during the computation because we assigned kernel value one instead of the K as traditional CNN \cite{zahan2022sdfa}. Figure \ref{fig:main_diagram}(b) visualizes the Sep-TCN, and Figure \ref{fig:Sep_TCN} demonstrates the details about the portion of the Sep-TCN where first extracted the local temporal proximity through the wider space. Figure \ref{fig:main_diagram}(b) demonstrates the Sep-TCN, including the DW, PW and residual connection, where it is included max-pooling for enhancing the temporal feature. It mainly focuses on the dominant frame as a feature vector.

\subsection{Architecture for Branches of Spatial-Temporal Approach } \label{sec4.3}
We previously mentioned that embedding the skeleton data with their corresponding position simultaneously provides a latent space that can be considered learning-specific feature information. After extracting the Sep-TCN feature, we applied three streams of the spatial-temporal branch to convey the corresponding private feature separately. We extracted the hand skeleton dataset from the original images using a media pipe where each image generated N joints for a single time frame. Each joint consists of a K-dimensional vector where K=3 (X, Y, and Z). For detailed understanding, we assume $g$ is the number of images, and after extracting skeleton points, it produced the T × N × K where t represents the time, N, K defines the number of joints and the number of channels or dimensionality. In our case, we consider T=4, N=21 and K=3.

\begin{figure}[H]
\begin{adjustwidth}{-3cm}{-3cm}
\centering
\centering 
\includegraphics[scale=0.35]{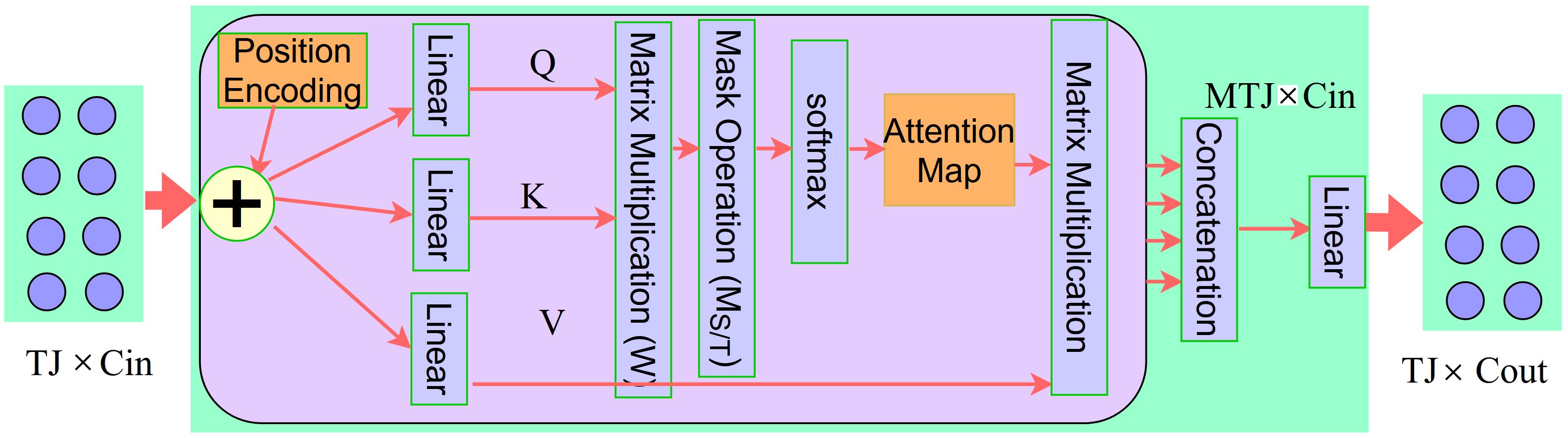}
\caption{Internal Structure of Multihead Attention Mechanism}
\label{fig:Attention}
\end{adjustwidth}
\end{figure}

The first branch included one stage, spatial attention with spatial embedding, which investigates and discovers the relationship between the adjacent joints within a specific time frame. In the second stream, we included an attention model with temporal embedding where the main purpose of the second stream is to investigate and discover the interaction between the neighbour joint within more than one frame. The third branch sequentially combined two stages, namely spatial attention and temporal attention, where the first stage produced the spatial features and the second stage produced the temporal features. The third branch spatial attention stage's main goal is to produce the spatial feature, then encode and discover the adjacent joint's interaction with a certain frame. As a consequence, it can produce the fine-grained joint-level features of the specific gestures g. Figure \ref{fig:Attention} showed the inner structure of the attention and masking operation.

\subsubsection{Spatial-temporal position embedding} \label{sec4.4}
There are three streams of the spatial-temporal attention model where we encoded positional embedding to maintain the sequence of the joint information because there are no built-in functions in the attention model to keep the skeleton data in sequences. We can consider each gesture skeleton joint set as a data tensor to feed the attention-based spatial-temporal neural network. There is no predefined structure or order for this skeleton node to visualize the identity of the node, as a consequence, it is not possible to identify the node or gesture name. To solve a specific node's identity crisis problem, we propose a spatial, temporal position embedding approach for generating the information based on the joint information. We performed the spatial-temporal position embedding by adding the extra dimension to the original skeleton dataset. Here, spatial embedding is a vector containing the position of all joints in a single frame. On the other hand, temporal embedding is a vector which consists of the position of the sequence of the frame in single pieces. To calculate the position embedding, we employed sine and cosine functions based on the different frequencies for embedding the position number for every skeleton node using the following Equation~\cite{shi2020decoupled,chen2019construct,miah2023dynamic}. 

\begin{equation} \label{Eq_PosEmb_1} 
\begin{array}{l}
P_{E}(p,2i)=sin\left(p/1000^{2i/C_{in}} \right)\\
P_{E}(p,2i+1)=cos\left(p/1000^{2i/C_{in}} \right)
\end{array}
\end{equation}

Here i and p defined the position of each individual element,  the sin function we employed to calculate the position embedding for the odd index, and the cos function we used to calculate the position for the odd index. After calculating the position embedding, we added the position embedding with the initial feature vector generated by the Sep-TCN network. Consequently, the embedded value of the feature vector associated with the corresponding position identity value is generated. As we employed here, three different modules of the spatial-temporal approach produced the three different kinds of vectors, which are defined by the following formulas: 

\begin{equation} \label{Eq:PosEmb_2} 
\bar{f}_{ST(t,i)}=A_{S}\left(f_{(t,i)}+P^{S}_{(t,i)} \right)
\end{equation}
\begin{equation} \label{Eq_PosEmb_3} 
\bar{f}_{TS(t,i)}= A_{T}\left(f_{(t,i)}+P^{T}_{(t,i)} \right)
\end{equation}      
\begin{equation} \label{Eq:PosEmb_4} 
\bar{f}_{ST(t,i)}=A_{T}\left(P^{T}_{t,i}+A_{S}\left(f_{(t,i)}+P^{S}_{(t,i)} \right) \right)
\end{equation}

Here, Equation~(\ref{Eq:PosEmb_2}),(\ref{Eq:PosEmb_2}),(\ref{Eq:PosEmb_2}) produces the final feature $f\bar{f}_{ST(t,i)}$, $bar{f}_{TS(t,i)}$ and $bar{f}_{ST(t,i)}$ for spatial, temporal and spatial-temporal for a specific initial feature node $v_{(t,i)}$ consequences. In each Equation $f_{(t,i)}$ defined the initial feature which is generated by the Sep-TCN, Temporal and spatial attention approach is defined by the  $A_{T}$, $A_{S}$ respectively. For a specific frame of the hand gesture is t and i-th number joint, that frame is defined here with the $P^{S}_{(t,i)}$ and $P^{T}_{(t,i)}$ the output dimension where the embedding dimension is the same as the input $f_{(t,i)}$ dimension. 

\subsubsection{Multihead Attention Module} \label{sec4.5}
In the study, we proposed three streams of attention models that included three different attention approaches which consisting of spatial, temporal and temporal-spatial attention branches. Each attention branch is combined with the position embedding for the spatial, temporal and spatial-temporal branches. Each branch took the input from the output of the Sep-TCN, and then the first and second branches calculated the spatial attention feature and temporal attention features. The third branch included two stages; the first stage took the input from the Sep-TCN output and then updated the initial feature with the spatial attention block; after that, fed the spatial feature into the temporal feature for producing the spatial-temporal feature vector. Multhihead attention we employed in each branch to perform the attention mechanism \cite{chen2019construct,miah2023dynamic,shi2020decoupled}. Figure \ref{fig:Attention} showed the inner structure of the attention and masking operation. Assume the initial separable feature is $f_{(t,i)}$ for a specific joint node V of a specific frame g which fed to the attention mechanism for the attention layer. To perform single-head attention, we first calculate the fully connected layer to project the query, key and value vector with a separable input feature $f_{(t,i)}$. The projection mechanism is performed with the following Equations (\ref{Eq:multihead_1}):

\begin{equation} \label{Eq:multihead_1} 
Q^{m}_{(t,i)}=W^{m}_{Q}f_{(t,i)},K^{m}_{(t,i)}=W^{m}_{K}f_{(t,i)}, V^{m}_{(t,i)}=W^{m}_{V}f_{(t,i)}
\end{equation}

Where, $Q^{m}_{(t,i)}$, $K^{m}_{(t,i)}$, and $V^{m}_{(t,i)}$ are defined the query, key and value nodes, respectively. The weight matrix for the m-the head attention can be denoted by $W^{m}_{Q}$, $W^{m}_{K}$, $W^{m}_{V}$ and this weight matrix is computed from the three stages; the first stage computes the dot product between the first two matrices and applies the softmax activation to normalize the output. In the second stage, we employed the masking operation to determine the type of attention, either spatial or temporal domain. This is mainly decided by assigning some value to the weight matrix. For example, if we determine spatial attention, we block the temporal position by assigning 0 and pass the spatial value by assigning one and vice versa \cite{chen2019construct,miah2023dynamic,shi2020decoupled}. In the third stage, we multiplied the spatial or temporal feature with the value vector and produced the feature for the m-th head attention. In the same way, we computed the attention for eight heads and concatenated to produce the combined feature for the multi-head using the following Equation (\ref{Eq:multihead_2}). 

 \begin{equation} \label{Eq:multihead_2} 
\bar{f}_{(t,i)}=Concat[\bar{f}^{1}_{(t,i)},\bar{f}^{2}_{(t,i)},\bar{f}^{3}_{(t,i)},...,\bar{f}^{M}_{t,i}]
\end{equation}

The total number is defined here with the M, and we consider 8 in our study. Our model produced the spatial feature $A_{S}$  from the 1st branch, the temporal feature $A_{T}$ from the 2nd branch and the spatial-temporal $A_{ST}$ feature from the 3rd branch. 

\subsubsection{Mask Operation. } \label{sec4.6}

In our study, to reduce the computational complexity, we first employed the Sep-TCN to minimize the parameters of the model and mask operation to reduce the computational complexity of the attention model sharply. In each stream of the architecture, we employed masking operations to cut down the computational cost based on spatial and temporal concepts. In the mask operation, we assign 0 and 1 based on the type of attention block where 0 means to block or cut down the dimension based on value and one means to pass the value. For the spatial attention module, we assign 1 for all the values for the spatial position and 0 for the temporal position, which means it blocks all the values for the temporal block and, consequently, produces the spatial features. In the same procedure we followed for the temporal attention block, we assigned 1 for the temporal position for passing the temporal feature and blocked the spatial feature by assigning 0 to the spatial position. The mask operation yields the reduced feature, which also sharply reduces the data block size, making the model faster and more efficient by reducing the computational cost~\cite{chen2019construct,miah2023dynamic,shi2020decoupled}. In short, we can say that masking operations make Multihead Self-Attention (MHSA) either a spatial or temporal attention model. The internal mechanism for the masking operation is demonstrated in Figure \ref{fi:Mask_Operation}. 

\begin{figure}[H]
\begin{adjustwidth}{-3cm}{-3cm}
\centering
\centering 
\includegraphics[scale=0.40]{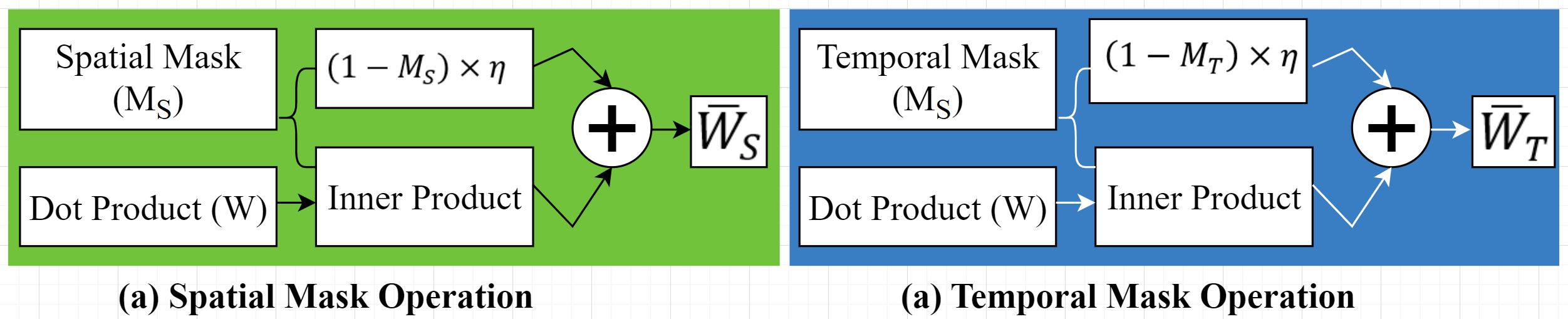}
\caption{Internal Structure of the masking operation}
\label{fig:Mask_Operation}
\end{adjustwidth}
\end{figure}


\subsection{Classification} \label{sec4.7}
After concatenating three stream features, we made a final feature vector, which we fed into the classifier module. After that, we applied a classification module shown in Figure \ref{fig:main_diagram}(c). Which included a fully connected layer, rectified linear activation, normalization, leaky ReLU activation and dropout layer. The fully connected layer has 128 input features and 128 output features. This rectified linear activation function applies element-wise ReLU to the input tensor. The normalization layer applies layer normalization to the input tensor. Layer normalization normalizes the features across the channel dimension for each sample in the batch. After that, a leaky ReLU activation function with a negative slope 0.1. It applies element-wise leaky ReLU to the input tensor. Where used is an argument that indicates that the operation will modify the input tensor directly without allocating additional memory. The dropout layer randomly zeroes some of the elements of the input tensor with a probability value. The overall purpose of this sequence of layers is to transform the input features into a higher-level representation that is better suited for classification. The layers use nonlinear transformations and normalization techniques to extract meaningful features from the input. The dropout layer helps prevent overfitting by randomly dropping out some input features during training. 
In our model, we averaged the video to make the matrix into a vector and applied an activation layer.
As the averaging technique specifically, we used the "row-wise averaging" technique in our implementation. This technique calculates the mean or average value of each row in the data. This operation computes the average of each row in the com array that we fed into the activation layer.
For simplicity, we finally employed a categorical cross-entropy during training \cite{miah2022bensignnet}, Equation(\ref{Eq:classification_1}) showed the formula of loss and prediction calculation.

     \begin{equation} \label{Eq:classification_1} 
Loss(\varnothing,\hat{\varnothing}) = -\sum_{c=1}^C \varnothing_c \log \hat{\varnothing}_c
\end{equation}
Here, C is the number of classes, $\varnothing$ is the ground truth, and $\hat{\varnothing}$is the prediction. 

\subsection{Dimension of Feature Map}
In our research study, we leveraged the Hand Pose Media Pipe system to extract a set of 213 skeleton points for each frame. Here, the number '21' represents the total count of joints in the hand, while '3' corresponds to the x, y, and z coordinate values associated with each of these skeletal points. For each individual trial within our dataset, we carefully structured it to encompass three frames, with the intent of creating a sequential or consecutive frame sequence within each trial. To effectively manage and process our data, we further organized these individual trials into batches, with each batch consisting of 8 trials. This bundling of data led to the dimensional representation of a single batch being the dimension becomes $Batch \times Trial$, where Batch is 8 here. 
This dataset, with its specified dimensions, was then seamlessly integrated into the Sequential Temporal Convolutional Network (Sep TCN) architecture. The Sep TCN, adept at handling sequential data, transformed our input data into a higher-level representation. Consequently, the output from the Sep TCN exhibited a dimension of $8\times 128 \times 3\times 21$. Here, the '128' signifies the number of output channels, '3' denotes the number of frames per trial, '21' represents the number of joints, and '8' corresponds to the total number of batches. Then we fed this feature dimension into four distinct streams: (i) Spatial Attention Map with Spatial PosEmbedding (ii) Temporal Attention map with Temporal Position Embedding.
(iii) Spatial Attention Map with Spatial Position Embedding-Temporal Attention Map with Temporal Position Embedding (iv) Residual Connection. Each of these streams independently generated feature representations characterized by dimensions of $8\times 84\times 128$. Subsequently, to consolidate the information from these diverse streams, we concatenated the feature representations from all four streams. This fusion resulted in a unified feature representation with dimensions of $8\times 336\times 128$. This concatenated feature representation was then forwarded to the classification module, which played a pivotal role in mapping the extracted features to their respective class labels. After this classification step, we applied an averaging technique, specifically referred to as "row-wise averaging." This technique computed the mean along the rows of our feature representation, thereby reducing the dimensionality to $8 \times 128$. Finally, this refined feature representation was channelled into the fully connected layer, which ultimately yielded the final output corresponding to the distinct classes in our classification task. The detailed description of the dimension of a feature map process outlines the essential steps involved in our research methodology, demonstrating the meticulous handling and transformation of input data as it traverses through our model architecture to ultimately yield the desired classification outcomes.

\section{Experimental Results} \label{sec5}
We tuned the parameters in each machine-learning model. This section demonstrated the evaluation metrics, parameter tuning, experimental setting, results and comparison with the state-of-the-art model. After that, the reason for the distance- and angle-based features are discussed. 

\begin{table}[h]
\caption{\textcolor{blue}{Possible hyperparameters}}
\label{Tab:hyparameters}
\begin{tabular}{lll}
\hline
Hyperparameter Name & Proposed Model & Existing Transfer Learning \\
\hline
Training : Testing & 70\%:30\% & 70\%:30\% \\ 
Dp rate & 0.01 & 0.01 \\ 
Learning Rate & 5e-6  to 1e-3 & 1e-3 \\ 
Optimizer & Adam & Adam \\ 
Batch Size & 8 - 32 & 8 \\ 
Epochs & 100-1000 & 500 \\
Patient & 100 & 100 \\ \hline
\end{tabular}
\end{table}

\subsection{Environmental Setting} \label{sec5.1}
To evaluate our proposed model when split into training and testing like 70\% and 30\%, respectively. In addition, to prove the generalization property, we consider the inter, intra, and merge evaluation settings. Intra-dataset settings define the training and testing that come from the same dataset. Inter dataset is that one dataset is used for training, and a different dataset is used for testing. To implement the model, we used PyTorch python framework and Google Colab Pro edition environment \cite{abadi2016tensorflow}. This was mainly developed with Tesla P100, which integrated 25GB GPU~\cite{tock2019google}. To implement the concept of computational Graph compatibility and adaptability with minimum resources and open-source properties pytorch has become a boon to the deep learning model. For the initial preprocessing of the image, we use OpenCV python package~\cite{gollapudi2019opencv} and Mediapipe to extract the hand key points. After extracting the hand key points, we make it a CSV file and a pickle file. To process the mathematical and statistical operation of the dataset, we used Numpy and Pandas Python packages which mainly give us various facilities for the various matrix operations. To visualize the various figures, we used the Matplotlib package. Table \ref{Tab:hyparameters} The table reveals several critical hyperparameters used in the experiment.  The first column contains the hyperparameter name, the second column represents the proposed model, and the third column represents the existing Transfer Learning-based model we experimented on here for validation purposes. In the hyperparameter column, first, the "dp rate" represents the dropout rate, typically set at 0.01 to regulate overfitting. The "learning rate" determines the optimizer's step size, typically within the Adam optimizer. Although not explicitly defined, "batch size" is crucial for training, usually set when creating data loaders. The "epochs" denote the number of training iterations, with 500 specified in this context. Additionally, "patience" controls early stopping, though it lacks an explicit definition. While these parameters are evident, it's essential to acknowledge that there may be other hyperparameters tied to model architecture, data preprocessing, or optimization that are relevant but not explicitly detailed in the provided code.
As a training parameter, we epochs 100-1000 and initial learning $5\mathrm{e}{-6} to 1\mathrm{e}{-3}$, which considered the higher fluctuation during the optimizer like ADAM and Nesterov  momentum~\cite{glorot2010understanding,dozat2016incorporating}. Moreover, in this study, we employed different parameter tuning operations to optimise the learning rate and optimizer with multiclass. 

\subsection{Evaluation metrics} \label{sec5.2}
Measuring the performance of the proposed model for the BSL recognition, we used accuracy, precision, recall and F1-score, which are mainly computed from the false positive (FP), true positive (TP), false negative (FN)  and true negative(TN) \cite{miah2022bensignnet}. The calculation formula for calculating the performance matrix is given below. Accuracy is calculated according to Equation (\ref{Eq:accuracy}), where the total correctly predicted classes are divided by the total count.

 \begin{equation}\label{Eq:accuracy} 
               Accuracy=(TP+TN)/(TP+FN+FP+TN) \times 100      
               \end{equation}
Precision is calculated according to Equation (\ref{Eq:precision}), where predicted positive sign classes are divided by the total positive count.

                \begin{equation} \label{Eq:precision} 
                         Precision=TP/(TP+FP)\times 100    
                          \end{equation}  
 The recall is calculated according to Equation (\ref{Eq:recall}), where predicted positive sign classes are divided by the total count of true positive and false negative.
 
  \begin{equation} \label{Eq:recall} 
                         Recall=T_p/(T_p+F_n )\times 100                                      \end{equation}
                         
F1-score is calculated according to Equation (\ref{Eq:f1_score}), where predicted twice of precision and recall multiplication divided by the summation of the precision and recall.

 \begin{equation} \label{Eq:f1_score} 
F1-score=(2 \times Precision \times Recall)/(Precsion+Recall)         
 \end{equation}
 \subsection{Result Analysis} \label{sec5.3}
We evaluated the model with inter-, intra, and merge dataset evaluation settings to generate the generalisation property. We arranged seven evaluation configurations here to measure the performance of the proposed model, as shown in Table \ref{Tab:Performance_Configuration}. In Table \ref{Tab:Performance_Configuration}, the intra-dataset evaluation setting is defined in configurations A, B, and C, which means the training and testing dataset comes from the same dataset. Inter-dataset evaluation settings are denoted with the configurations D and E, which refer to the training and test datasets from different datasets. For instance, in configuration D, the model is trained and validated on the Proposed dataset and tested with the BdSL-38 dataset; in configuration, the E model is trained and validated with the Proposed dataset before being tested on the BdSL 38 dataset. Merge dataset validation settings are defined with the configuration F and G. This refers to the fact that we train the model by merging the BdSL-38 and BAUST-BSL-38 datasets and testing with BdSL-38 and BAUST-BSL-38 datasets, respectively. 

\begin{table}[h]
\caption{Performance accuracy with different Dataset configuration}
\label{Tab:Performance_Configuration}

\begin{tabular}{lllll}
\hline
Configuration & Train   Dataset & Test   Dataset & Performance   {[}\%{]} & Category \\ \hline
Configuration   A & KU   BdSL & KU   BdSL & 99.00 & Intra   dataset \\ 
Configuration   B & BAUST-BSL-38 & BAUST-BSL-38 & 98.00 & Intra   dataset \\
Configuration   C & BdSL38 & BdSL38 & 96.00 & Intra   dataset \\
Configuration   D & BAUST-BSL-38 & BdSL38 & 32.60 & Inter   dataset \\
Configuration   E & BdSL38   & BAUST-BSL-38 & 31.60 & Inter   dataset \\
Configuration   F & \begin{tabular}[c]{@{}l@{}}BdSL38+\\ BAUST-BSL-38\end{tabular} & BAUST-BSL-38 & 95.00 & Merge   dataset \\
Configuration   G & \begin{tabular}[c]{@{}l@{}}BdSL38+   \\ BAUST-BSL-38\end{tabular} & BdSL38 & 92.00 & Merge   dataset \\ 
\hline
\end{tabular}
\end{table}

\begin{figure}[H]
\begin{adjustwidth}{-3.5cm}{-3cm}
\centering
\centering 
\includegraphics[scale=0.33]{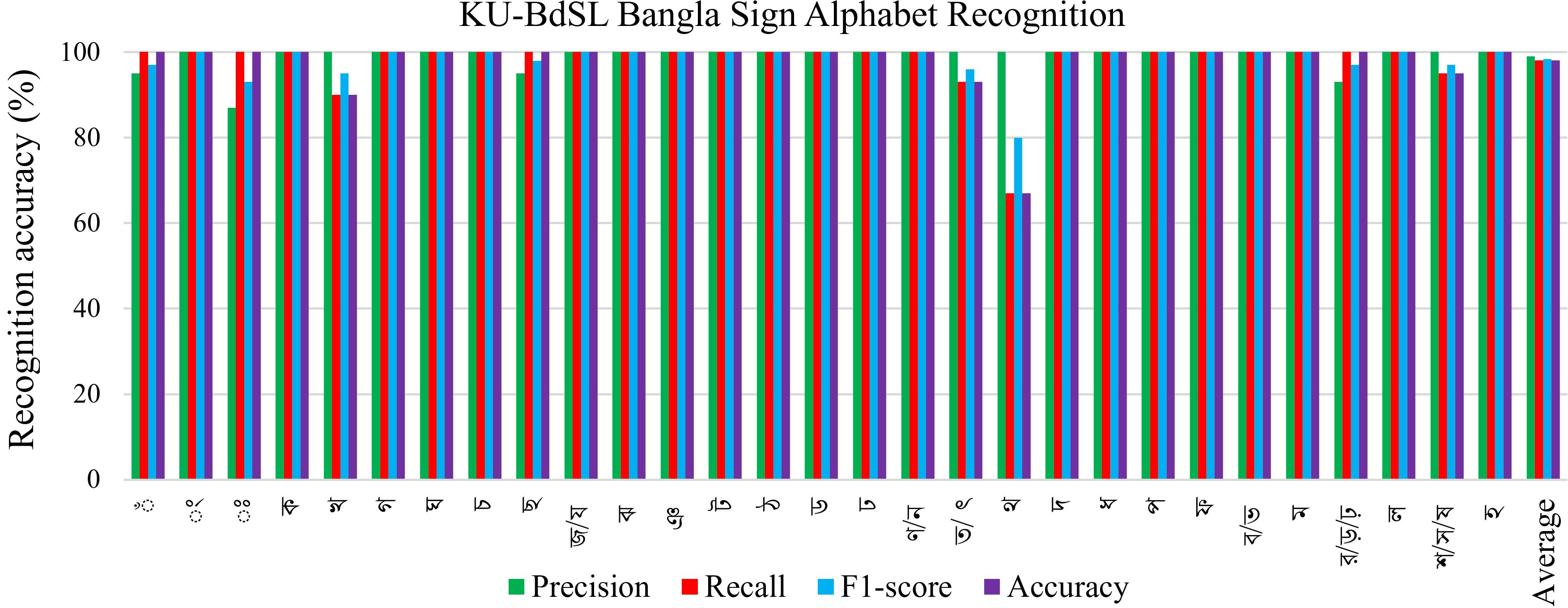}
\caption{Class wise Precision, Recall, F1-score and accuracy of the KU BdSL Dataset Configuration A}
\label{fig:kubdsl_performance}
\end{adjustwidth}
\end{figure}

\begin{figure}[H]
\begin{adjustwidth}{-3cm}{-3cm}
\centering
\centering 
\includegraphics[scale=0.35]{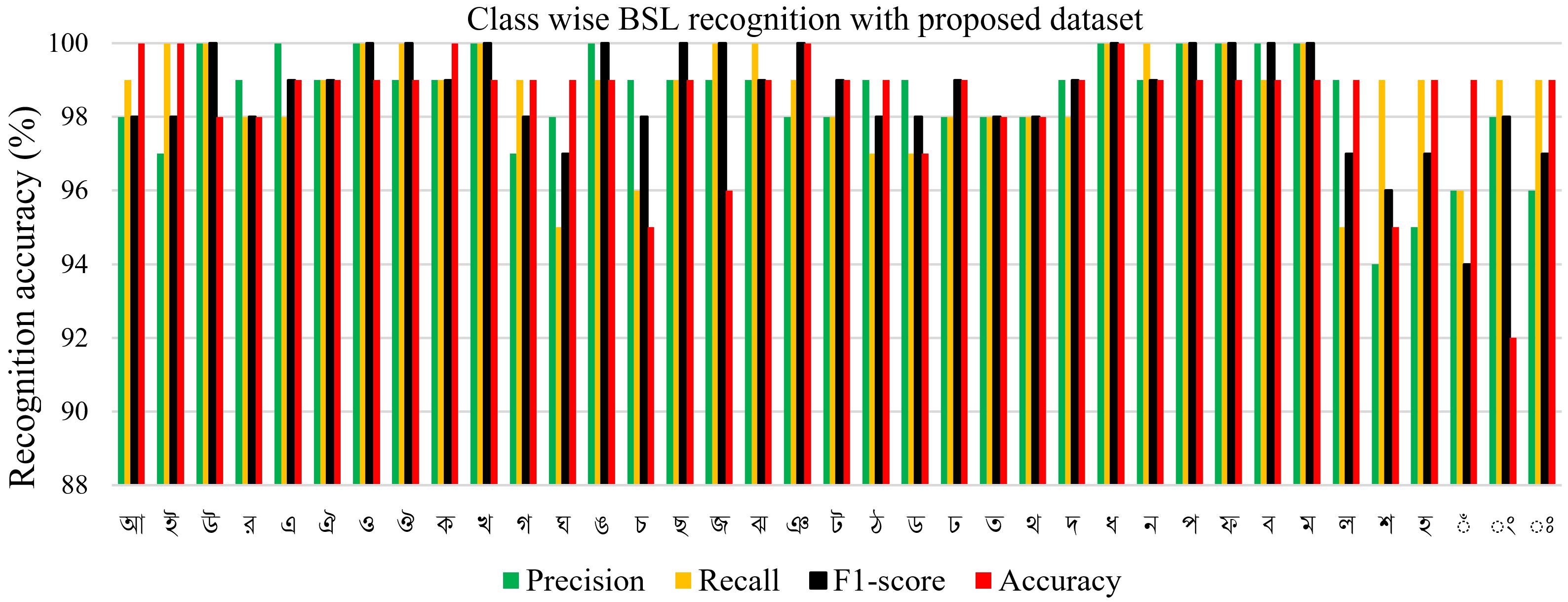}
\caption{Class wise Precision, Recall, F1-score and accuracy of the proposed dataset configuration B}
\label{fig:baust_bdsl_performance}
\end{adjustwidth}
\end{figure} 

\begin{figure}[H]
\begin{adjustwidth}{-3cm}{-3cm}
\centering
\centering 
\includegraphics[scale=0.30]{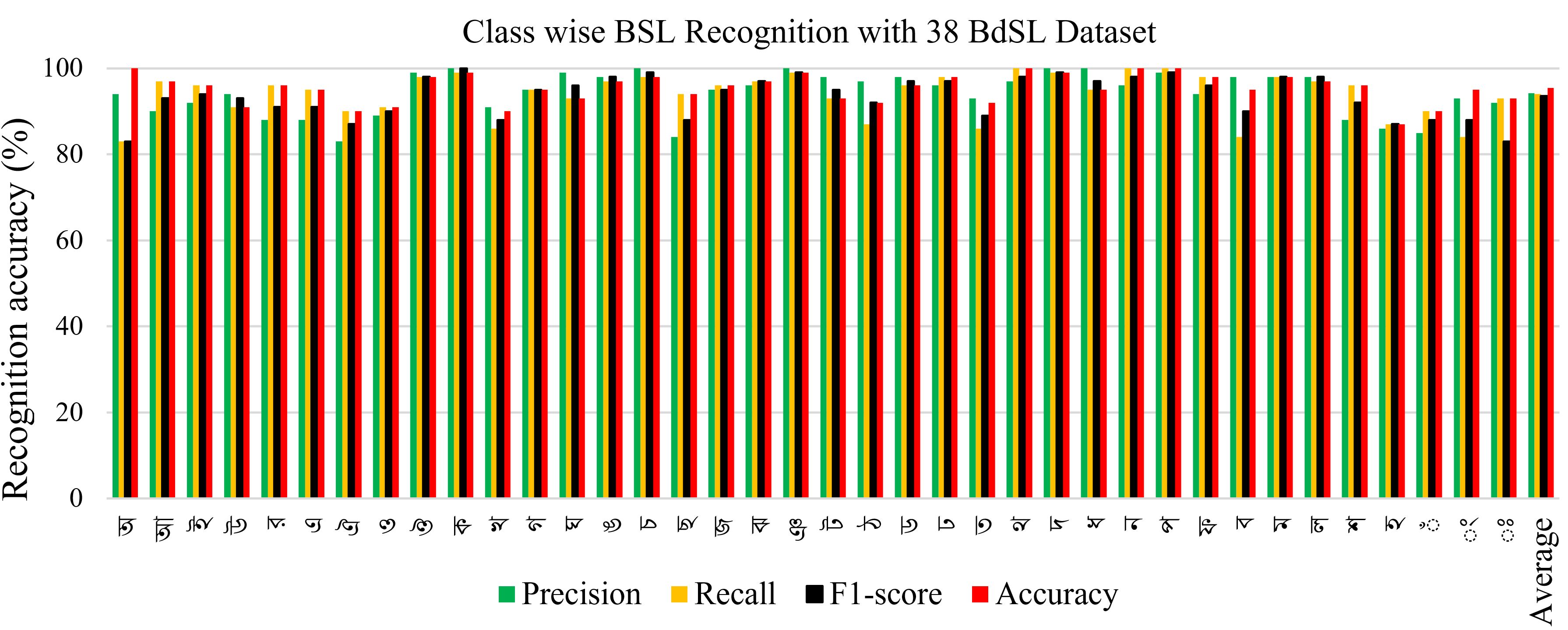}
\caption{Class wise Precision, Recall, F1-score and accuracy of the BdSL 38 dataset Configuration C. }
\label{fig:BdSL-38}
\end{adjustwidth}
\end{figure} 

We have performed a thorough assessment of the proposed models trained on different configurations of inter-datasets and intra-datasets to determine the generalizability of the proposed model. In our work, we have considered the training set from one dataset to evaluate the trained model. Furthermore, we have tested the trained model with different dataset test sets and highlighted the generalization problem. We demonstrate that the validation partition is useful in the intra-dataset as it is drawn from the same dataset as the training set. The results in Table \ref{Tab:Performance_Configuration} show that the accuracy of all configurations trained on inter-datasets, which had previously performed well on intra-datasets, has significantly decreased. We demonstrated label-wise precision, recall, f1-score and performance accuracy for the intra-dataset in Figures \ref{fig:kubdsl_performance},\ref{fig:baust_bdsl_performance}, and \ref{fig:BdSL-38} for the KU-BdSL, Proposed, and BdSL-38 datasets, respectively. 

\begin{figure}[H]
\begin{adjustwidth}{-3cm}{-3cm}
\centering
\centering 
\includegraphics[scale=0.30]{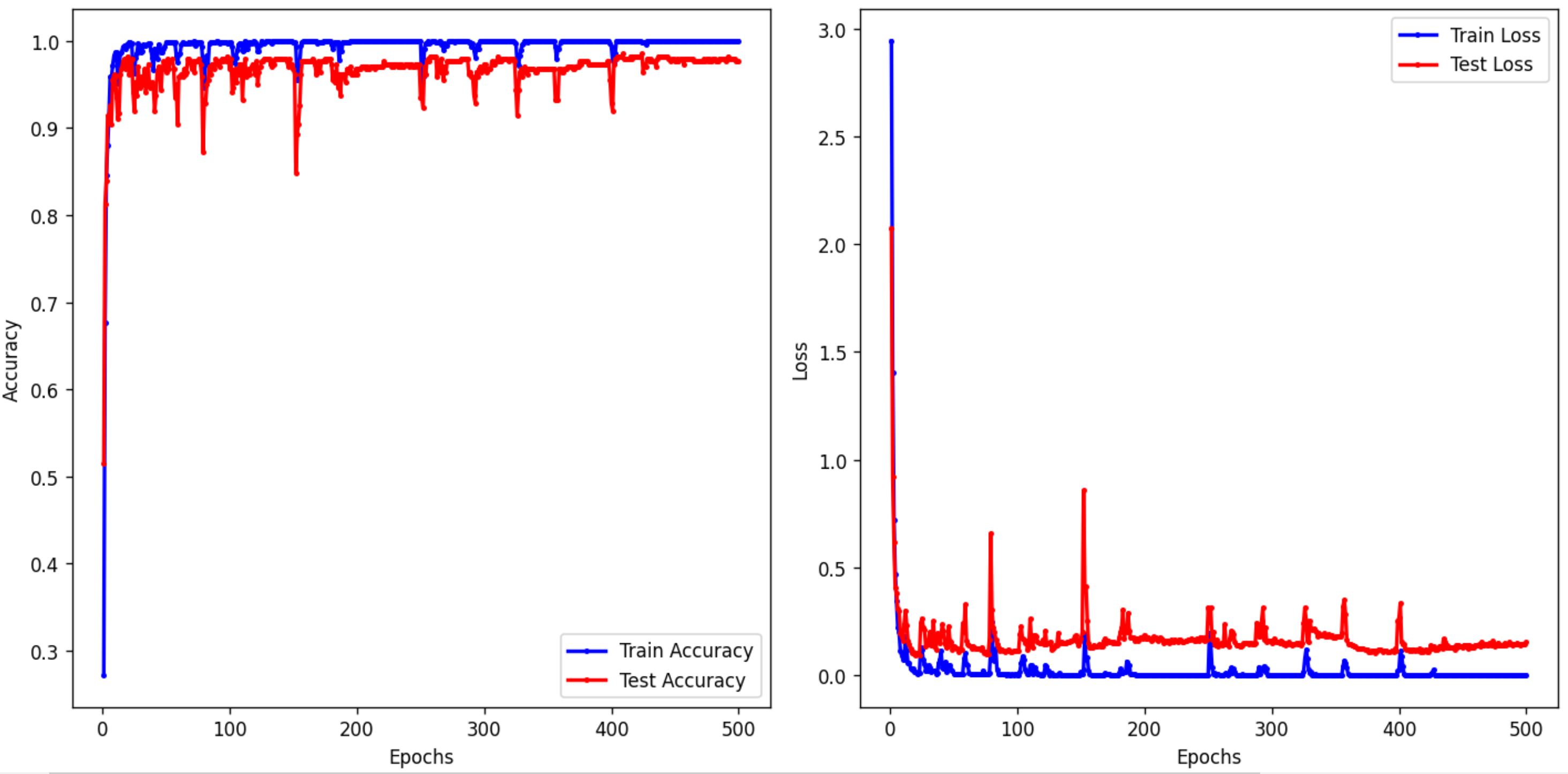}
\caption{Accuracy and Loss curve for the configuration A. }
\label{fig:curve_KuBd}
\end{adjustwidth}
\end{figure} 

Figure \ref{fig:curve_KuBd} illustrates the relationship between training accuracy versus testing accuracy and training loss versus testing loss, focusing specifically on configuration A among various configurations. The figure portrays a sample curve that provides insights into the model's fit within the context of our learning algorithm's objectives. It effectively strikes a balance between overfitting and underfitting. In our experience, an ideal fit is characterized by both training and validation losses diminishing until they reach a stable point, with minimal disparity between their final values. Typically, the model's loss on the training dataset is lower than that on the validation dataset, leading to an anticipated gap between their respective learning curves. This gap is commonly referred to as the "generalization gap." Our learning curves exemplify a favourable fit, where the plot of training loss steadily decreases to a point of stability, mirroring the behaviour of the validation loss plot, which also converges to a stable value with a small, acceptable gap when compared to the training loss.

\subsection{State of the Art Comparison of the proposed model} \label{sec5.4}
\begin{figure}[H]
\begin{adjustwidth}{-3cm}{-3cm}
\centering
\centering 
\includegraphics[scale=0.35]{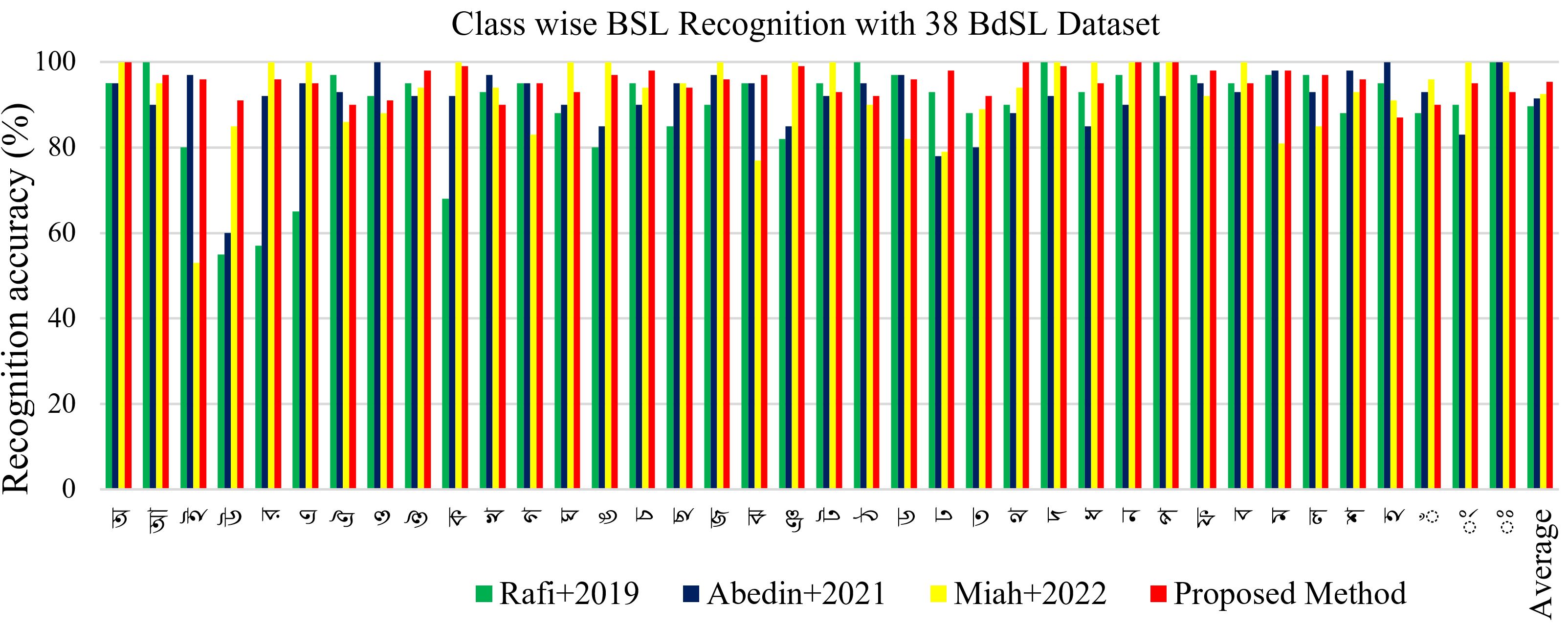}
\caption{Label-wise state-of-the-art comparison for BdSL-38 dataset.}
\label{fig:Comparison_BdSL-38}
\end{adjustwidth}
\end{figure}

Table \ref{Tab:comparison_BdSL-38} demonstrated the comparative analysis of the proposed model with the state-of-the-art system for the BdSL-38 dataset, where we reported the number of parameters and computational time also. Our proposed model has an order of magnitude lower computational complexity (in terms of FLOPS) and also has fewer (1/2nd) parameters compared to existing architectures. This reduces hardware requirements in terms of processing power and memory, leading to lower costs. Reduced computational complexity also results in faster run time, as shown in Tables \ref{Tab:comparison_BdSL-38} and \ref{Tab:comparison_KU-BdSL}. Our method takes less time, about three times than the best-competing method. 
It can be observed from Table \ref{Tab:comparison_BdSL-38} that the proposed method has reported the best results in test accuracy when compared to the other models. In \cite{rafi2019image}, the author employed modified VGG19 to recognize Bengali sign language alphabets. With the learning rate 1e-3 and SGD optimizer, they achieved 89.60\% accuracy. In \cite{abedin2021bangla}, the author employed a concatenation CNN architecture where they concatenated the feature extracted from the Bengali sign language dataset and achieved 91.52\% accuracy. To implement their concatenation network, they used 10 ReLU activations, a single input layer, four max-pooling layers, two fully connected layers and a softmax activation function. The author in \cite{miah2022bensignnet} applied a unique CNN-based architecture, namely BenSignNet, to recognize the Bengali sign language dataset, and they achieved 94.00\% accuracy where, whereas our proposed model achieved 96.00\% accuracy. Apart from performance, it is clear that DenseNet, with 92.29\%, outperforms the other model, whereas ResNet performs worse than all for the BdSL-38 dataset. Figure \ref{fig:Comparison_BdSL-38} visualizes the label-wise performance comparison with the state-of-the-art models for the BdSL-38 dataset.
\begin{table}[h]
\caption{Comparison with state-the-state of the art model for BdSL38 Dataset}
\label{Tab:comparison_BdSL-38}
\begin{tabular}{lllll}
\hline
Dataset                  & Approach             & Parameter (M)   & Flops (BMac)       & Accuracy (\%) \\  \hline
\multirow{8}{*}{BdSL 38} & VGG19{\cite{rafi2019image,simonyan2014very}}     & N/A         & N/A         & 89.60         \\
                         & ResNet \cite{he2016deep}      & 11.68   &  248015   & 89.68         \\
                         & InceptionNet \cite{szegedy2016rethinking} & 47.82       & N/A       & 88.00         \\
                         & CNN \cite{abedin2021bangla}         & N/A         & N/A         & 91.52         \\
                         & WaveNet   \cite{hao2021human}   & N/A         & N/A         & 92.00         \\
                         & DenseNet \cite{mohameed2021automated}    & 0.676       & N/A         & 92.29         \\
                         & BenSignNet \cite{miah2022bensignnet}  &  1.42    & 33088 & 94.00         \\
                         & Proposed Method            & 0.26639  & 210   & 96.00  \\
                         \hline
\end{tabular}
\end{table}

\begin{table}[h]
\begin{adjustwidth}{-2cm}{-3cm}
\caption{State-of-the-art comparison of the proposed method for KU-BdSL dataset}
\label{Tab:comparison_KU-BdSL}
\begin{tabular}{lllll}
\hline
Dataset           & Approach              & Parameter (M) & Flops (BMac)         & Accuracy (\%) \\  \hline
\multirow{2}{*}{} & BenSignNet \cite{miah2022bensignnet}   & 1.42 M    & 1537.37   & 98.20         \\
                  & InceptionNet \cite{szegedy2016rethinking} & 47.82     & N/A           & 94.94         \\
KU-BdSL           & WaveNet \cite{hao2021human}      & N/A       & N/A           & 90.00         \\
\multirow{3}{*}{} & ResNet \cite{he2016deep}      & 11.68   &  11525& 98.21         \\
                  & DenseNet \cite{mohameed2021automated}     & 0.676     & N/A           & 98.21         \\
                  & Proposed Model        & 0.26639 M &  9.78       & 99.00   \\    
                  \hline
\end{tabular}
\end{adjustwidth}
\end{table}

\begin{figure}[H]
\begin{adjustwidth}{-3cm}{-3cm}
\centering
\centering 
\includegraphics[scale=0.35]{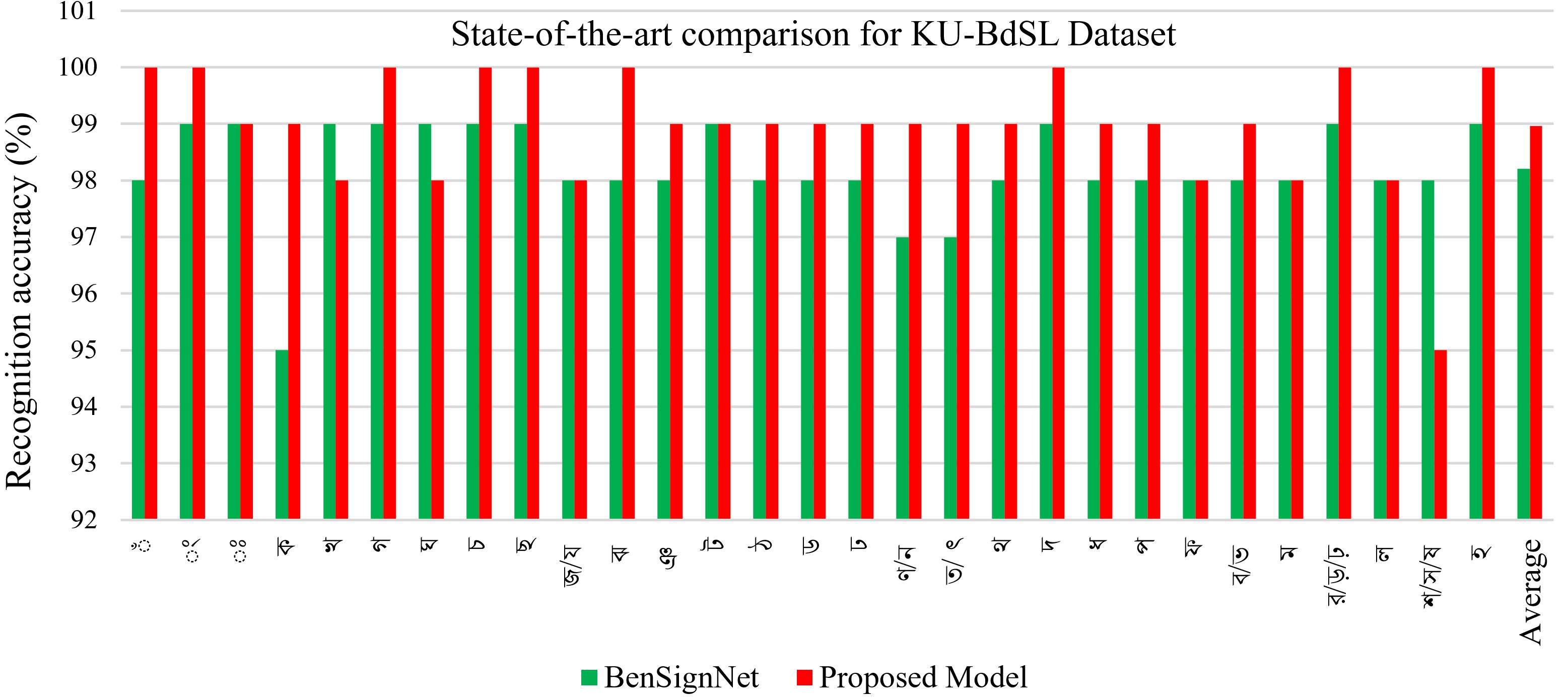}
\caption{Label-wise state-of-the-art comparison for KU-BdSL dataset.}
\label{fig:Comparison_KU-BdSL}
\end{adjustwidth}
\end{figure} 
Tables \ref{Tab:comparison_KU-BdSL} demonstrated the comparative analysis of the proposed model with the state-of-the-art system KU-BdSL dataset, where we reported the number of parameters and computational time also. Table \ref{Tab:comparison_KU-BdSL} visualized that the proposed model achieved 99.00\% accuracy, which is better than all other existing models. 
\begin{table}[h]
\begin{adjustwidth}{-1cm}{-3cm}
\caption{State-of-the-art comparison for the newly  created BAUST-BSL-38 lab dataset}
\label{Tab:comparison_Lab_Dataset}
\begin{tabular}{lllll}
\hline
Dataset & Approach & Parameter & Flops & Accuracy (\%) \\ \hline
\multirow{6}{*}{BAUST-BSL-38  Lab Dataset} & VGG16\cite{simonyan2014very} & 14.50 & 415.17 & 79.00 \\
 & InceptionNet \cite{szegedy2016rethinking} & 47.82 & N/A & 80.00 \\
 & WaveNet \cite{hao2021human} & N/A & N/A & 78.00 \\
 & DenseNet \cite{mohameed2021automated} & 0.676 & N/A & 97.07 \\
 & BenSignNet \cite{miah2022bensignnet}  & 1.37 & 32749 & 78.00 \\
 & ResNet \cite{he2016deep}      & 11.68   &  245474& 96.77         \\
 & Proposed & 0.26639   M & 208.10 & 98.00  \\ \hline
\end{tabular}
\end{adjustwidth}
\end{table}

We also demonstrate here the performance of different transfer learning model performance. Based on the Experiment WaveNet \cite{hao2021human}, InceptionNet \cite{szegedy2016rethinking}, ResNet \cite{he2016deep} and DenseNet~\cite{mohameed2021automated} achieved 90.00\%,  94.94\%, 98.21\%, and 98.21\% accuracy, respectively. Meanwhile, DenseNet and ResNet achieved the nearest performance for the KUBdSL dataset. Figure \ref{fig:Comparison_KU-BdSL} visualizes the label-wise performance comparison with the state-of-the-art models for the KUBdSL.

Table \ref{Tab:comparison_Lab_Dataset} showcases the performance of different state-of-the-art models on the BAUST-BSL-38 lab dataset. VGG16, with 14.50 million parameters, achieves a 79\% accuracy, consuming 415.17 million FLOPs. In contrast, InceptionNet, despite a high parameter count of 47.82 million, reaches 80\% accuracy and WaveNet lags slightly with a 78\% accuracy. DenseNet impresses with a 97.07\% accuracy, owing to a relatively low parameter footprint. BenSignNet, with 1.37 million parameters and 32,749 FLOPs, also scores a 78\% accuracy. ResNet generated 96.77\% accuracy accuracy but with higher parameters and FLOPs. The proposed model stands out with a stellar 98\% accuracy, the lowest parameter count at 0.26639 million, and moderate FLOPs at 208.10 million, showcasing a balance of efficiency and high performance.


\subsection{Discussion}

Our proposed Multi-Branch Spatial-Temporal Network model, depicted in Figure \ref{fig:main_diagram}, has achieved high-performance accuracy while maintaining low computational complexity. This significant reduction in computational demands leads to an order of magnitude lower complexity in FLOPs and features fewer parameters—just half the count of existing architectures. Table \ref{Tab:comparison_BdSL-38} details the proposed model, which requires only 0.266 million parameters per batch, with each batch comprising eight trials for each dataset. It also demands 210 billion multiply-accumulate operations (BMac) as computational complexity for 2,733 batches. In comparison, the number of parameters for ResNet, Inception, BenSignNet, and DenseNet are 11.68, 47.82, 1.42, and 0.602 million, respectively. Notably, our model's parameter count is approximately one-fourth that of BenSignNet and half of DenseNet's, also falling below those of ResNet and Inception models. DenseNet has the lowest parameter count among existing systems at 0.602 million—roughly double that of our proposed model at 0.266 million. In terms of computational complexity measured in FLOPs, our proposed model requires 210 BMac, while methods like ResNet and BenSignNet need 248,015 and 33,088, respectively. BenSignNet has the lowest complexity at 33,088, which is still about three times higher than our proposed model's 210 BMac. According to Table \ref{Tab:comparison_KU-BdSL}, the number of parameters is consistent with Table \ref{Tab:comparison_BdSL-38}, but computational complexity varies based on the number of batches. The KU-BdSL dataset, with 127 batches, shows our model requiring only 9.78 BMac, while methods like ResNet and BenSignNet need 11,525 and 1,537, respectively. BenSignNet's lowest complexity is still roughly three times higher than our proposed model's 9.78 BMac. In Table \ref{Tab:comparison_Lab_Dataset}, the parameter number remains consistent with Tables \ref{Tab:comparison_BdSL-38} and \ref{Tab:comparison_KU-BdSL}, but again, computational complexity varies by batch count. For the BAUST-BSL-38 lab dataset's 2,705 batches, our model requires just 208.10 BMac, compared to the significantly higher demands of ResNet and BenSignNet, which require 245,474 and 32,749, respectively. BenSignNet, with the lowest complexity of 32,749, is more than three times higher than that of our proposed method at 208.10 BMac.

The proposed model marks a substantial leap in the spatial-temporal model, striking an impressive balance between efficiency and accuracy. With a parameter count of merely 0.26639 million, it vastly undercuts the parameters of models like VGG16 and InceptionNet by an order of magnitude. This streamlined architecture directly translates into reduced memory consumption and quicker training times, ideal for deployment in resource-constrained environments. Moreover, the computational cost, in terms of FLOPs, is about a third of that required by previous models such as VGG16 and ResNet. This decrease in computational demand, without a sacrifice in performance, underscores the innovative and efficient nature of our model. It not only positions the model as more environmentally friendly due to lower energy consumption but also paves the way for real-time applications on edge devices with limited computational resources. The model's commendable 98\% accuracy rate, despite a reduced architecture, establishes a new standard for future developments, demonstrating that high efficiency does not necessitate a trade-off in performance. This reduction in hardware requirements for processing power and memory also leads to cost savings. Lower computational complexity equates to faster runtimes, as evidenced in Tables \ref{Tab:comparison_BdSL-38},\ref{Tab:comparison_KU-BdSL}, and \ref{Tab:comparison_Lab_Dataset} with our method taking approximately one-third of the time compared to the next best-competing method. Our model, which uses 2D skeletons extracted from images taken with standard cameras and does not rely on power-intensive and costlier active sensors like the Kinect, shows that hand gesture recognition can be effectively generalized across different conditions. To validate this generalization, we assessed performance across multiple datasets under varying conditions and in both inter- and intra-evaluation settings. While most existing studies report high accuracy on one or two datasets, our use of three datasets underscores the model's robustness and generalizability, consistently handling a variety of gestures.

\section{Conclusion}

In this study, we introduced a lightweight BSL recognition system utilizing a spatiotemporal attention approach. Our system effectively detects BSL from 2D skeleton data, which eliminates the need for video storage or costly sensors. We have developed a potent feature aggregation technique that discerns local and global spatiotemporal attributes, leading to high accuracy in our system. The exceptional accuracy of our Bangla Sign Language recognition model stems from a unique multi-branch spatial-temporal attention mechanism that integrates four distinct and effective feature types: (i) spatial attention from the first branch, (ii) temporal attention from the second, (iii) combined spatial-temporal attention from the third, and (iv) temporal CNN features from the fourth. This innovative fusion approach is pioneering, utilizing the full range of informative cues for superior gesture classification. By fusing these features, our model captures the nuances of BSL more effectively than existing methods. Its computational efficiency and ability to integrate diverse datasets highlight its adaptability and potential for real-time application. This approach marks a significant leap forward in sign language recognition technology. The proposed architecture benefits from low computational complexity and precise graph convolution attention. Our extensive experiments across three BSL datasets have demonstrated that our model surpasses existing systems in various settings and evaluations. The results section visualizes the model's strong performance in intra-dataset and merged dataset configurations, although it reveals less promising results in inter-dataset settings. The study also indicates that while the current datasets are sufficiently large for training the model for recognition, there could be gaps in representing diverse scenarios. Our model possesses a generalizable quality, making it suitable for deployment in real-world applications to aid the disabled community. However, the model may encounter issues, such as underfitting with smaller datasets like KU-BdSL and Ishara-Lipi etc. In the future, we plan to compile a large-scale BSL dataset encompassing a variety of scenarios to train the model, ensuring its continued high performance across diverse situations.

\section{Sample Appendix Section}
\label{sec:sample:appendix}

 \bibliographystyle{elsarticle-num} 
 \bibliography{cas-refs}





\end{document}